\documentclass[accepted]{article}

\usepackage{microtype}
\usepackage{graphicx}
\usepackage{subcaption}
\usepackage{booktabs} % for professional tables

\usepackage{hyperref}

\usepackage{icml2026}

\usepackage{amsmath}
\usepackage{amssymb}
\usepackage{mathtools}
\usepackage{amsthm}

\usepackage[capitalize,noabbrev]{cleveref}

\usepackage{tikz, tikz-cd, graphics, graphicx, mathtools, amsfonts, mathrsfs, amsmath, xfrac, dsfont, enumitem, multirow, mathdots, tabularx, wrapfig, tabularray}

\newcommand{\bigO}[1]{\mathcal{O}{\left( {#1} \right)}}

\newcommand{\reals}{\mathbb{R}}

\newcommand{\abs}[1]{\left\lvert#1\right\rvert}
\newcommand{\norm}[1]{\left\lVert#1\right\rVert}

\theoremstyle{plain}
\newtheorem{theorem}{Theorem}[section]

\theoremstyle{definition}

\theoremstyle{remark}
\newtheorem{remark}[theorem]{Remark}

\usepackage[textsize=tiny]{todonotes}

\icmltitlerunning{SSM Adapters via Hankel Reduced-order Modeling}

\begin{document}

\twocolumn[
  \icmltitle{SSM Adapters via Hankel Reduced-order Modeling: Injection Site Determines Task Suitability in Long-Context Fine-Tuning}
  \icmlsetsymbol{equal}{*}

  \begin{icmlauthorlist}
    \icmlauthor{Omanshu Thapliyal}{HAL}
  \end{icmlauthorlist}

  \icmlaffiliation{HAL}{Hitachi America Ltd., Santa Clara, CA, USA}

  \icmlcorrespondingauthor{Omanshu Thapliyal}{omanshu.thapliyal@hal.hitachi.com}

  \icmlkeywords{State Space Models, Controllability, Observability, Grammian, Hankel rank reduction}

  \vskip 0.3in
]



\printAffiliationsAndNotice{}  % no special notice (required even if empty)

\begin{abstract}
While parameter-efficient fine-tuning (PEFT) typically targets attention projectors, its efficacy for tasks requiring sequential state accumulation remains under-explored.
We examine if PEFT for such tasks can benefit from state space model (SSMs) adapters, and if MLP blocks are better injection sites.
We introduce Hankel Reduced order Model (HRM) adapter, an SSM-based residual module initialized via Balanced Truncation of empirical Hankel Grammians.
By leveraging the time-invariance of the system matrix $\bar{A}$, HRM enables an exact FFT-based parallel scan, achieving computational parity with LoRA across all context lengths.
In iso-parametric evaluations on Mistral-7B (8.4M trainable parameters), HRM outperforms LoRA variants on LongBench tasks, including QuALITY (+34.8\% relative accuracy) and QMSum (+71.6\% relative ROUGE-1).
HRM further demonstrates consistent superiority across 18 configurations of synthetic state-tracking (DFA, Parity) and character-level language modeling (enwik8).
Gate analysis reveals that HRM adapters effectively learn to modulate recurrence, providing a robust architectural alternative to low-rank adaptation for long-context sequence modeling.
% Abstracts must be a single paragraph, ideally between 4--6 sentences long.
% Gross violations will trigger corrections at the camera-ready phase.
\end{abstract}

\section{Introduction}
Parameter-efficient fine-tuning (PEFT) is a dominant paradigm in adapting large pre-trained language models (LLMs) for downstream tasks.
Rather than updating entire model weights, PEFT methods insert adapters or modify a smaller subset of parameters, keeping the model backbone frozen.
Low-Rank Adaptation \cite{hu2022lora} is the most widely adopted PEFT method, achieving strong results across language understanding, generation, and instruction following tasks while adding $\sim$0.1–1\% extra parameters.
LoRA parameterizes a weight update as $\Delta W=BA$, where $B\in\reals^{d_{out}\times r}$ and $A\in\reals^{r\times d_{in}}$.
The matrices $A,B$ are learned such that rank $r\ll \min{(d_{out},d_{in})}$, where the full forward pass through the adapted layer becomes:
\begin{equation}
h_t=W_0 x_t + BAx_t = (W_0+BA)x_t
\end{equation}
where the model weights $W_0$ are kept frozen for input $x_t$, at position $t$.
We observe that the computation to adapt weights in LoRA (and its related methods: DoRA \cite{liu2024dora}, QloRA \cite{dettmers2023qlora}, AdaLoRA \cite{zhang2023adalora}) is a static linear function of the input $x_t$.
As a result, the adapter output at position $t$ has no access to the prior positions: $x_{t-1}, x_{t-2},\cdots.$.
This is not a failure of reduced rank modeling as no choice of $r$ will give LoRA temporal memory access.

To motivate this central issue, consider fine-tuning a model to simulate a 4-state Deterministic Finite Automaton (DFA).
At each step, the correct output depends not on the current input symbol alone, but on the accumulated sequence of transitions since the start. 
A DFA with 4 states can be in any of 4 configurations depending on the entire history $x_1, x_2, \cdots, x_{t-1}$. 
LoRA, regardless of rank, collapses the current state to a static function of $x_t$, therefore, it structurally cannot represent a state that persists across positions.
Despite this, LoRA, DoRA, AdaLoRA, and QLoRA's successes in achieving excellent results on tasks where adaptation is position-independent, such as domain style transfer, factual knowledge injection, and instruction following is well established in literature.

To this end, we investigate the following question:
\emph{is it possible to construct a PEFT adapter that (1) adds temporal recurrent state to a frozen transformer, (2) is provably compressible to a minimal state dimension, (3) and is computationally equivalent to LoRA, while achieving better performance on long-range tasks across diverse domain?}

\begin{figure*}[!ht] % [t] for top placement is standard for figure*
  \centering
  \resizebox{\textwidth}{!}{%
    \input{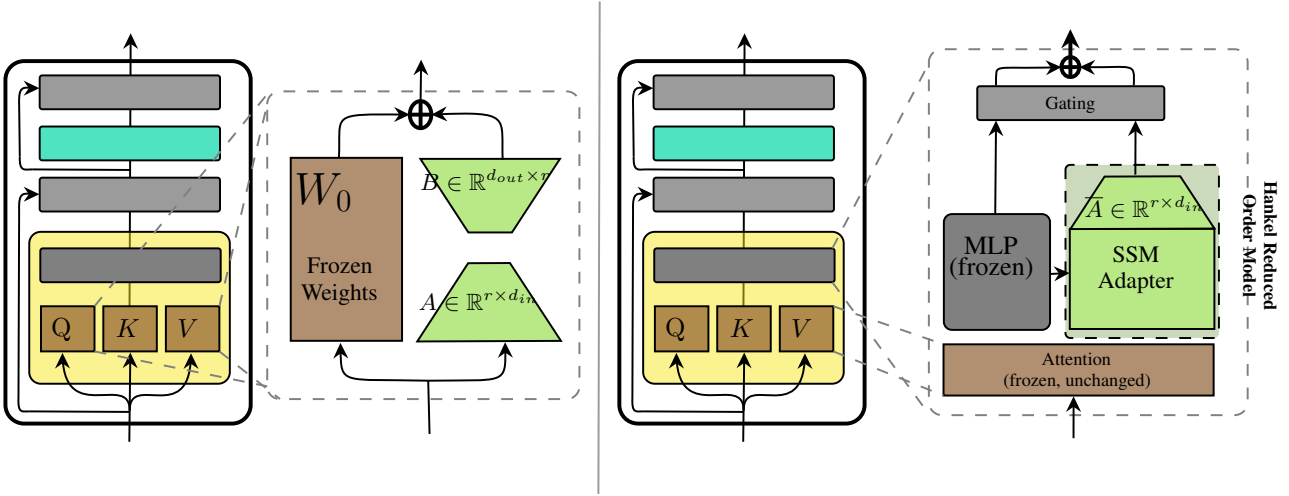}
  }
\caption{Architecture comparison. LoRA modifies weight matrices; its output at position $t$ is a static function of $x_t$. The HRM adapter inserts a parallel recurrent branch whose hidden state integrates all prior representations.}
\label{fig:lora-hrm}
\end{figure*}

\section{Related Works}
All major \textbf{PEFT} methods share the common structural property of position-independence (or position agnostic weight fine-tuning).
LoRA \cite{hu2022lora}, AdaLoRA \cite{zhang2023adalora}, QLoRA \cite{dettmers2023qlora}, LoRA+ \cite{hayou2024lora+}: all compute $h = f(x_t)$ with no dependence on t or prior positions. 
AdaLoRA adaptively allocates rank but the resulting update is still a static matrix product.
IA$^3$ \cite{liu2022few} applies learned vectors to rescale hidden states, a multiplication by a position-independent scalar, also resulting a static update (see Fig.~\ref{fig:lora-hrm}).

Some works prepend learned soft tokens to the input \cite{lester2021power, li2021prefix}.
These tokens provide context at the input but do not define a recurrent state, and the transformer still processes each position independently after the prefix.
Foundational adapter methods for PEFT \cite{houlsby2019parameter, pfeiffer2020adapterhub} insert small MLP bottlenecks to pre-trained transformer models.
The bottleneck $h = W_2 \cdot \sigma(W_1 \cdot x_t)$ depends only on $x_t$m without memory recurrence.
While \cite{houlsby2019parameter} places two adapter bottleneck modules into each transformer layer, while \cite{pfeiffer2020adapterhub} places a single adapter, halving the number of trainable parameters.

On the other hand, State Space models (SSMs) have been shown promise to alleviate the quadratic attention costs over long-contexts.
Structured State Space Sequence (S4) Models were introduced by \cite{gu2021efficiently} with state space layer with HiPPO-based initialization and convolution-mode inference, which was improved in S4D \cite{gu2022parameterization} by restricting to diagonal state space matrix $\bar{A}$, losing expressiveness but enabling simpler inference.
Finally, Mamba models \cite{gu2023mamba} introduced input-dependent state matrices $(A_t,B_t)$, enabling selective memory.

\textbf{Hybrid model architectures} such as Griffin \cite{de2024griffin}, MambaFormer \cite{park2024can}, and Jamba \cite{lieber2024jamba} utilize SSM layers with transformers.
On the surface they seem similar to our proposed work (HRM inserts a $d=32$ SSM at each MLP block; MambaFormer inserts Mamba layers between attention blocks). 
The critical distinction is one of training regime: every hybrid architecture requires joint training from scratch on billions of tokens. 
HRM is the first method that adds SSM-style temporal memory in the PEFT setting; therefore, the backbone is frozen, the adapter has $\sim0.1\%$ parameters, and no pre-training data beyond the fine-tuning task is required. 
As a result, a user with a frozen, pre-trained GPT-2 cannot apply MambaFormer to it, but they can apply HRM.

The combination of (a) recurrent hidden state, (b) provable compression via model-order reduction, and (c) computational parity with static adapters does not appear in the literature.
To the best of our knowledge, the closest related work is SLoRA (\cite{sheng2024slora}) and related low-rank SSM approaches that treat SSMs as a structured alternative to LoRA rank approximations. 
However, these do not apply Balanced Truncation, do not provide error bounds, and do not address the computational overhead of the recurrence.

\section{Background}
\paragraph{LoRA}
Low-Rank Adaptation (LoRA) \cite{hu2022lora} relies on the observation that weight updates during fine-tuning $\Delta W\in\reals^{d_{out}\times d_{in}}$ of pre-trained models lie in a low intrinsic dimension \cite{aghajanyan2021intrinsic}.
This motivates parameterizing the update as a rank-$r$ product:
\begin{equation}
\Delta W=BA, B\in\reals^{d_{out}\times r} \text{ and }A\in\reals^{r\times d_{in}}
\end{equation}
During training $W_0$ is frozen, and only $B$ and $A$ are updated. 
At inference, the update is absorbed as $W_{eff} = W_0 + BA$, adding no inference latency, with the forward pass:
\begin{equation}
h=W_{eff}x=(W_0+BA)x=W_0x+B(Ax)
\end{equation}
LoRA is applied to the $Q$ and $V$ projection matrices of each self-attention block in standard practice. For a model with $n_{layers}, d_{model}$ attention dimension, this contributes $4r \cdot n_{layers} \cdot d_{model}$ trainable parameters. 
The mapping $h=(W_0+BA)x$ is a linear function of $x$ alone. 
The matrix $W_{eff}$ is fixed at all positions. 
If we index the sequence position as $t$, the LoRA output at position $t$ is $h^{LoRA}_t=(W_0+BA)x_t$ with no dependence on the previous inputs $x_{t-1}$, etc.

The adapter applies the same linear transformation B A to every token, independently of position or context, and is therefore memory less.
AdaLoRA \cite{zhang2023adalora} addresses rank allocation but not memory either.
It parameterizes $\Delta W = P\Lambda Q$ where $P$, $Q$ are orthogonal and $\Lambda$ is diagonal (singular value decomposition structure), pruning entries of $\Lambda$ based on importance. 
The result is still a static linear map of the current token.
QLoRA \cite{dettmers2023qlora} addresses memory efficiency (4-bit quantization of $W0$) and DoRA \cite{liu2024dora} decomposes into magnitude and direction components.
Both remain static functions of the current token. 
The memory-less property is therefore preserved in existing LoRA variants.

\paragraph{SSMs}
A continuous-time linear state-space model (SSM) is defined by the equations:
\begin{equation}\label{eq:continuous-ssm}
\dot{x}(t)= Ax(t) + Bu(t), \;y(t)=Cx(t) + Du(t)
\end{equation}
for hidden state $x\in\reals^d$, input $u\in\reals^{m}$, output $y\in\reals^{p}$, and the state-transition (or system) matrix $A$, $B$ the input matrix, $C$ the output matrix, and $D$ the feed-through, or skip matrix.
For sequence modeling, the continuous-time system is discretized to obtain a recurrence relation. 
Given a time step $\Delta t$, the Zero-Order Hold (ZOH) discretization yields:
\begin{equation}\label{eq:discrete-ssm}
\begin{split}
x_{t} &= \bar{A}x_{t-1} + \bar{B}u_t, \;y_t=Cx_t\\
\bar{A} &=e^{A\Delta t}, \bar{B}=A^{-1}(e^{A\Delta t}-I)B  
\end{split}
\end{equation}

The discrete SSM defines a linear map from the input sequence $\{u_1, ..., u_T\}$ to the output sequence $\{y_1, ..., y_T\}$:
\begin{equation}\label{eq:ssm-convolution}
y_t= \sum^{t}_{k=0}\underbrace{C\bar{A}^{t-k}\bar{B} }_{g_{t-k}} u_k=(g\star u)_t
\end{equation}
where $g_{k}$ is the impulse response of the system.
This results in the output sequence be written as the causal convolution of the impulse response with the input.

Structured State Spaces (S4) \cite{gu2022parameterization} showed that when $\bar{A}$ is initialized as a specific Normal Plus Low-Rank (NPLR) matrix, the SSM can model long-range dependencies with a stable impulse response that decays slowly. 
The key computational insight of S4 is that the causal convolution $(g\star u)$ can be computed in $\bigO{(T \log{T})}$ via FFT.
We will also use this fact for our computation in the subsequent sections.

Finally, the \emph{stability of the discrete SSM} in (\ref{eq:discrete-ssm}) requires all eigenvalues of $\bar{A}$ to lie strictly within the unit circle, i.e., $\max{\abs{\lambda_i(\bar{A})}} < 1$. 
For diagonal $\bar{A}$ with real entries, this requires $\abs{\bar{A}_{ii}}<1$.
We will enforce this by parameterization, as our reduced-order modeling requires stability of the underlying linear time invariant (LTI) system.

\paragraph{Balanced Truncation in LTI Systems}
Consider LTI dynamics (i.e., fixed $G\triangleq(\bar{A},\bar{B},C,D)$) in (\ref{eq:discrete-ssm}), with state dimension $d$.
The reduced order modeling problem for LTI dynamical system $G$ is then to find a reduced order system $\hat{G}$ with state dimensions $\hat{d}<d$, such that the input-output behaviors of $G$ and $\hat{G}$ are as close as possible, with a quantified error bound.
\emph{Balanced Truncation} (BT) \cite{moore2003principal} is the canonical solution to this problem for stable LTI systems.

The LTI system's state $v\in\reals^d$ is \emph{controllable} if there exists an input sequence to drive $G$ from the origin to $v$.
Controllability for the LTI system relies on the \emph{Controllability Grammian} $W_c \in\reals^{d\times d}$, a positive semi-definite matrix defined as:
\begin{equation}\label{eq:ctrb}
W_c=\sum^\infty_{k=0} \bar{A}^k\bar{B}\bar{B}^T(\bar{A})^k
\end{equation}
Equivalently, $W_c$ is known to be the solution of the discrete time Lyapunov equation \cite{corless2003linear}:
\begin{equation}\label{eq:ctrb-lyap}
\bar{A}W_c\bar{A}^T-W_c+\bar{B}\bar{B}^T=0
\end{equation}

Conversely, the state $v$ is \emph{observable} if the initial state $v$ can be uniquely determined from the output sequence $\{y_k\}$.
Similarly, observability for the LTI system relies on its \emph{Observability Grammian} $W_o\in\reals^{d\times d}$, defined as:
\begin{equation}\label{eq:obsv}
W_o=\sum^\infty_{k=0} (\bar{A}^T)^kC^T C \bar{A}^k
\end{equation}
with its corresponding Lyapunov equation:
\begin{equation}\label{eq:obsv-lyap}
\bar{A}^TW_o\bar{A}-W_o + C^TC=0
\end{equation}

Matrices $W_c$ and $W_o$ play an important role in balanced truncation of $G$ by forming a joint Hankel operator $\mathcal H:\text{\{past inputs\}} \to \text{\{future outputs\}}$.
For discrete time LTI system, $\mathcal H$ is the fixed matrix $\Gamma\triangleq W_cW_o$, which maps the full causal history of inputs to all future outputs. 
To perform truncation, we need to align the coordinate system so that directions are ordered by their joint controllability/observability.
The diagonal entries $\sigma_i$'s are called the \emph{Hankel singular values} (HSVs):
\begin{equation}\label{eq:HSV}
\sigma_i=\sqrt{\lambda_i(W_cW_o)},\;\;\sigma_1\geq \sigma_2\geq\cdots\geq \sigma_d\geq 0
\end{equation}
also the $i^\text{th}$ singular value of the Hankel operator, characterizing a state direction that is irrelevant to the past-to-future input-output map.
The balancing transformation is a coordinate transform $T\in\reals^{d\times d}$ such that the Grammians are simultaneously diagonalized:
\begin{equation}\label{eq:BT}
TW_cT^T=T^{-T}W_oT^{-1}=\Sigma = \text{diag}(\sigma_1,\cdots,\sigma_d)
\end{equation}
As a result, the transformed system $(T\bar{A}T^{-1},T\bar{B},CT^{-1})$ has the property that each state direction has equal controllability and observability, equal to $\sigma_i$.
Such a system is called a balanced system.

A balanced truncation $\hat{G}$ of the system $G$ can now be formed by partitioning the balanced system into ``important" ($1,\cdots,\hat{d}$) and ``unimportant" blocks ($\hat{d}+1,\cdots,d$) as:
\begin{equation}\label{eq:}\begin{split}
\hat{A} &= \left(T_{[1:\hat{d},1:\hat{d}]}\bar{A}T_{[1:\hat{d},:]}^{-1}\right)_{[1:\hat{d},1:\hat{d}]},\\
\hat{B} &=\left(T\bar{B} \right)_{[1:\hat{d},:]}, \hat{C} = \left(CT^{-1}\right)_{[:,1:\hat{d}]}
\end{split}
\end{equation}
Finally, Glover's error bound \cite{glover1984all} dictates that the truncated system deviates from the original by at most twice the sum of the discarded HSVs:
\begin{equation}
\norm{G-\hat{G}}_{\mathcal H_\infty} \leq 2\sum^d_{k=\hat{d}+1}\sigma_k
\end{equation}
This is a worst case bound over all inputs and all frequencies.
Furthermore, $\hat{G}$ is stable, and Glover's bound is tight.

\section{Method: Hankel-Reduced order Model Adapter}

\paragraph{Empirical Grammians for SSMs}
HSV-based balanced truncation requires that system $G$ be LTI.
The proposed HRM adapter's $\bar{A}$ is time-invariant, so the theorem applies directly. 
However, the selective SSM extension (input-dependent $B_t, C_t$ in selective scan \cite{gu2023mamba}) violates the LTI assumption. 
We address this extension via \emph{empirical Grammians} approach \cite{lall1999empirical}.
This extends balanced truncation to time-varying systems by approximation Grammians from observed state trajectories.
This involves running the system forward on a representative calibration dataset of $N$ sequences.
At each time step $t$ of each sequence $n$, we record the state vectors $s_{n,t}$, to compute the empirical controllability Grammian as:
\begin{equation}
W^{emp}_c=\frac{1}{N\cdot t_{cal}}\sum^{N}_{n=1} \sum^{t_{cal}}_{t=1} s_{n,t}s_{n,t}^T\in\reals^{d\times d}
\end{equation}
This estimates the covariance of state trajectories under typical inputs, a proxy for controllability.
A similar proxy for observability is found in the form of empirical observability Grammian as:
\begin{equation}
W^{emp}_o=\frac{1}{N\cdot t_{cal}}\sum^{N}_{n=1} \sum^{t_{cal}}_{t=1} y_{n,t}^Ty_{n,t}\in\reals^{d\times d}
\end{equation}
Due to \cite{lall1999empirical}, $W^{emp}_c\to W_c$ and $W^{emp}_o\to W_o$ as $N\to\infty$, and the converges at $\bigO{1/\sqrt{N}}$.

For our case, this gives a $\bigO{N\cdot T_{cal}\cdot d^2}$ to compute HSVs, from which the balancing transform and truncation proceed exactly as in the LTI case.
However, for the time-invariant HRM adapter, both the analytical Lyapunov and empirical Grammian approaches are available.

\paragraph{HRM Adapter Architecture}
Now we are ready to architect the HRM adapter based on Hankel order-reduction for our SSM.
Consider a standard pre-norm transformer layer $l$ with input $x_t \in \reals^{d_{model}}$. 
The layer applies self-attention followed by an MLP sublayer, each with residual connections and layer normalization:
\begin{equation}\label{eq:pre-trained-model-layer}
\begin{split}
a_t &=x_t+\mathrm{Attm}(\mathrm{LayerNorm}(x_t)),\\
h^{MLP}_t &=a_t+\mathrm{MLP}(\mathrm{LN}(a_t))    
\end{split}
\end{equation}
All weights Attn($\bullet$) and MLP($\bullet$) are frozen during adapter training.
The HRM adapter is inserted parallel to the $l^{\text{th}}$ MLP sub-layer, adding a recurrent correction to the MLP output:
\begin{equation}\label{eq:hrm-insert}
h^{out,(l)}_t=h^{MLP,(l)}_t + \alpha^{(l)}\cdot y^{(l)}_t
\end{equation}
where $\alpha^{(l)} \in \reals$ is a layer-specific learnable gate scalar and $y_t^{(l)}$ is the adapter output for layer $l$, at position $t$.

The adapter at layer $l$ defines a recurrent hidden state $s_t^{(l)} \in \reals^d$ that integrates the token representations $h_1, h_2, \cdots, h_t$ as they are processed:
\begin{equation}\label{eq:ssm-adapter-recurrence}
\begin{split}
s_t^{(l)} &= \bar{A}s_{t-1}^{(l)} + \bar{B}^{(l)} h_t^{MLP},\;\;y_t^{(l)}=C^{(l)}s_t^{(l)} \in \reals^{d_{domel}}\\
\end{split}
\end{equation}
where $\bar{A} \in \reals^{d\times d}$ is a learnable (diagonal) state transition matrix, $\bar{B} \in \reals^{d\times d_{model}}$ maps the current hidden state into the adapter's state space, $C \in \reals^{d_{model}\times d}$ maps the adapter state back to the hidden representation, and $\alpha$ is a learnable scalar gate.
Using (\ref{eq:hrm-insert}) and (\ref{eq:ssm-adapter-recurrence}) the adapter's output can be unrolled as:
\begin{equation}\label{eq:hrm-unrolled-state}
y_t^{(l)} = C^{(l)}\sum_{k=0}^t \left(\bar{A}^{(l)}\right)^{t-k} \bar{B}^{(l)} h^{MLP,(l)}_k
\end{equation}
Therefore, the combined gated addition (\ref{eq:hrm-insert}) gives the unrolled layer computation as:
\begin{equation}
h^{out,(l)}_t=h^{MLP,(l)}_t + \alpha^{(l)} C^{(l)}\sum_{k=0}^t \left(\bar{A}^{(l)}\right)^{t-k} \bar{B}^{(l)} h^{MLP,(l)}_k
\end{equation}

The adapter is placed parallel to the MLP, with a learnable scalar weight $\alpha$.
There are two reasons to this.
First, the attention mechanism already computes a weighted sum over all past positions, providing global context. 
Adding a recurrent branch to attention would interact with the causal mask in a non-trivial way and could disturb the attention distribution. 
Second, the MLP sub-layer is the natural site of position-independent computation as it applies the same learned function to each token representation independently. 
A recurrent adapter at this site adds the missing dependence on prior positions.
On the other hand, a sequential insertion would mean the MLP receives adapter-modified input, potentially causing large gradient flows through the frozen MLP. 
The parallel insertion ensures the frozen MLP is always evaluated on the original attention output, while the adapter's contribution is additive and controlled by $\alpha$.
The gating scalar $\alpha$ is initialized to a small value to ensure that at the start of the training the HRM adapter contribution is small, and the model starts from the behavior of a pre-trained backbone.
The gate then proceeds to grow with training and the adapter learns a useful temporal correction.
Experiments show that initializing at $\alpha_0=1.0$ causes divergence on all tested configurations.

Another design choice we made was to have time-invariant $\bar{A}$.
A natural concern is that Mamba-style selective SSMs (with input-dependent $\bar{A}_t, B_t, C_t$) are more expressive as they can selectively forget irrelevant tokens by adjusting the state decay on the fly. 
We fix in our architecture a time-invariant $\bar{A}$ deliberately because the pre-trained model's attention already handles selectivity to a certain degree.
The frozen self-attention mechanism performs global, content-based retrieval at every layer, choosing which past tokens to attend to. 
The HRM adapter's role is complementary: it provides a continuous recurrent state that integrates the local MLP output stream, accumulating context that attention's position-independent MLP stream cannot represent. 
Further, time-invariant $\bar{A}$ makes $\bar{A}^k$ a geometric sequence, enabling the exact FFT convolution shortcut
This eliminates the compute overhead and makes the HRM adapter practical, and temporal causal. 
Time-invariance also makes the trained adapter an LTI system, for which Balanced Truncation with the Glover $\mathcal H_\infty$ bound applies analytically. 
This also allows for an easier computation of Grammians using Lyapunov equation.
The expressivity trade-off is knowingly accepted in exchange for theoretical tractability and computational efficiency.

\paragraph{HRM State Transition Matrix Parameterization \& Stability}
We parameterize the state transition matrix $\bar{A}$ in (\ref{eq:ssm-adapter-recurrence}) as a fixed diagonal (this is common in Mamba and S4D like SSMs \cite{gu2023mamba, gu2022parameterization}) to keep matrix-vector product cost $\bigO{d}$, instead of $\bigO{d^2}$ for a general $\bar{A}$:
\begin{equation}
\bar{A}^{(l)} = \mathrm{diag}\left(\bar{a}^{(l))}_1,\cdots, \bar{a}^{(l))}_d\right),\bar{a}^{(l))}_i=\exp{(-\exp{(\log{A^{(l)}_i})})}
\end{equation}
where $\log{A^{(l)}_i}\in\reals$ is the raw (unconstrained) learnable parameter.
This parameterization is to ensure that $0<\bar{a}_i^{(l)}<1$, therefore, $\max_i{\abs{\lambda_i(\bar{A}^{(l)})}} < 1$ for each layer.
As a result, the HRM dynamics are unconditionally stable for any parameters values during all stages of training.

The parametrization above has a ZOH interpretation, as $\bar{A}=\exp{(A \Delta t)}$ for a continuous time system in (\ref{eq:continuous-ssm}) with ZOH discretization factor $\Delta t$.
The parameter $\bar{a}_i= \exp{(-\exp{(\log{A_i})})}$ corresponds to $\Delta t\cdot \abs{A_i}\in(0,\infty)$, with $\exp{(\log{A_i})}$ playing the role of $\Delta t \cdot \abs{A_i}$.
The combined parameter $\log A_i$ absorbs both, the magnitude or the continuous $A$ eigenvalue, and the discretization step $\Delta t$.
$\bar{B}$ and $C$ are unconstrained dense matrices with learnable parameters $\bar{B}^{(l)}\in \reals^{d\times d_{model}}$, and $C^{(l)}\in \reals^{d_{model}\times d}$, both initialized with small-variance Gaussian entries.
Finally, a learnable parameter $\log \Delta t^{(l)}$ associated with each layer $l$ is used to compute the ZOH discretization step $\Delta t^{(l)}$.
This allows the adapter to learn an appropriate timescale for the task.
As a result, the total number of learnable parameters per layer are: $B:d\times d_{model}$, $C: d_{model} \times d$, $\log A_i : d$, 
$\log \Delta t : d$,  and gate $\alpha: 1$ $= 2d\cdot d_{model}  + 2d + 1$ parameters.
Therefore, total number of parameters (compared with LoRA parameters) are:
\begin{equation}\label{eq:adapter-total-params}
\begin{split}
P_{HRM} &= n_{layers} \cdot ( 2d\cdot d_{model}  + 2d + 1)\\
P_{LoRA} &= n_{layers} \cdots 2\cdots2 \cdot r \cdot d_{model}
\end{split}
\end{equation}
Following this, a state compression is done so that for a fixed $r$, so that the iso-parametric $P_{HRM}$ is compressed for a Glover bound $\varepsilon=0.01$.
That is, top $90\%$ of the HSVs are kept, and the remaining discarded.
This HSV-based compression means that $P_{HRM}(\hat{d})\leq P_{HRM}\approx P_{LoRA}$.

\paragraph{Parallel Scan for HRM Adapter}
Since our $\bar{A}$ is time-invariant, the HRM recurrence computes a causal linear convolution. 
This convolution can be evaluated in $\bigO{T \log T}$ via the Fast Fourier Transform, replacing $\bigO{T}$ sequential Python-level dispatches with three FFT calls and achieving empirical compute parity with LoRA at all tested context lengths.
This parity is shown in the compute wall clock times for HRM and LoRA in Appendix.~\ref{app:fft-compute-time}.

Using FFT-based parallel scan arguments from \cite{gu2023mamba, gu2024mamba}, let $\{h_t\}_{t=0}^{T-1}\subset \reals^{d_{model}}$ be the input sequence to the HRM adapter.
Let $s_t$ and $y_t$ be the HRM sequential recurrence state and output, respectively, for $s_0=0$.
For the impulse response $g_k$ defined in (\ref{eq:discrete-ssm}), define the zero-padded sequences:
\begin{equation}
\begin{split}
\tilde{g}[k] &= \begin{cases} 
        g_k & 0\leq k\leq T-1 \\
        0   & T\leq k \leq 2T-1 
      \end{cases}\\
\tilde{h}[k]  &= \begin{cases} 
        h_k & 0\leq k\leq T-1 \\
        0   & T\leq k \leq 2T-1 
      \end{cases}
\end{split}
\end{equation}
Then the output sequence can be computed as:
\begin{equation}\label{eq:fft-output}
y_t = \sum^{t}_{k=0} t_{t-k}h_k=\left[\mathrm{IFFT}\left(\mathrm{FFT}(\tilde{g}\odot \mathrm{FFT}(\tilde{h}))\right)\right]_t
\end{equation}
for $0\leq t \leq T-1$, for element-wise product $\odot$, and FFT/IFFT operating on the length $2T$ dimensional sequence.
Since all operations in the FFT scan (\texttt{torch.fft.rfft}, element-wise multiply, \texttt{torch.fft.irfft}) are differentiable in PyTorch, gradients flow back through the FFT to $\bar{A}, \bar{B}, C$ without any custom CUDA kernels.

% ==========================================
\section{Experiments}
\subsection{Synthetic Task: DFA State Tracking}
Consider a deterministic finite automaton (DFA), a 5-tuple $(Q,\Sigma,\delta,q_0,F)$, where $Q$ is a finite set of states, $\Sigma$ an input alphabet, $\delta: Q \times \Sigma \to Q$ the transition function, $q_0$ the initial state, and $F\subseteq Q$ the set of accepting states. 
Suppose we are given an input sequence $(\sigma_1,\cdots,\sigma_T)\in \Sigma^T$, a DFA with $k$ states and binary alphabet $\Sigma = \{0, 1\}$.
The state tracking task is to predict the current DFA state at each position $t$, given $(\sigma_1,\cdots,\sigma_T)$, predict $q_t = \delta(q_{t-1}, \sigma_t)$.

This requires exact state accumulation, since $q_t$ depends on the full history $(\sigma_1,\cdots,\sigma_t)$ through the transition functions.
A model that cannot maintain state across positions will fail as $T$ grows as it must somehow compress the DFA state into the current token representation alone.

We experiment with $k\in\{2,4,8\}$, binary alphabet, context lengths $T\in\{64, 128, 256, 512\}$, with each DFA instance having a random fixed transition table $\delta$.
The model must output a $k$-way classification at each position, with sequences sampled uniformly at random from all valid DFA paths, 10,000 training sequences (1,000 validation sequences), across 3 different seeds.

\begin{figure}[!htbp]
    \centering
    \includegraphics[width=1\linewidth]{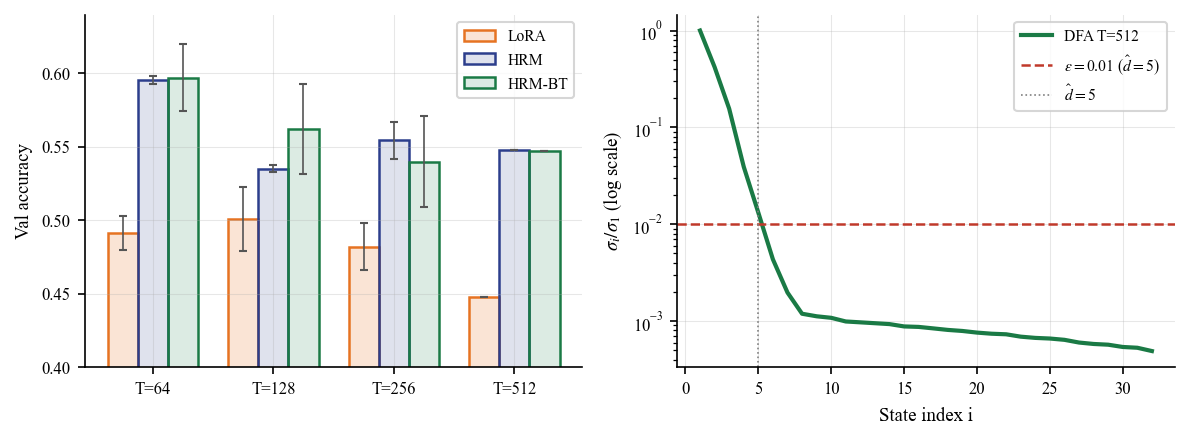}
    \caption{DFA state tracking results: (left) HRM vs. HRM with balanced truncation vs. LoRA, (right) Hankel Singular Value decay rate for the task, with HSV cutoff threshold = 0.01}
    \label{fig:dfa-result}
\end{figure}

In Fig.~\ref{fig:dfa-result}, we observe that HRM-BT dominates LoRA at all T values, and the gap grows with T, consistent with the memory hypothesis. 
Additionally, balanced truncation outperforms no truncation due to BT regularization. 
The HSV decay curve $\sigma_i/\sigma_1$ drops below 0.01 by $i=7$, justifying $\hat d$=6. 
The HSV spectrum confirms that \emph{DFA dynamics are intrinsically 6-dimensional, despite training with d=32 state dimensions}.

\section{MAESTRO Piano Language Modeling}
MAESTRO v2 \cite{hawthorne2018enabling} is a dataset of $\sim$200 hours of professional piano performances in symbolic MIDI format. 
We treat it as a character-level language modeling task: each MIDI event (note-on, note-off, time-shift, velocity) is encoded as a single token, and the model is trained to predict the next token given the context. 
The vocabulary has $\sim$300 distinct event tokens.
Piano music is an ideal testbed for long-range temporal modeling: (1) melodic phrases span dozens to hundreds of notes; (2) harmonic progressions follow conventions (ii-V-I, etc.) that span 8–16 measures; (3) rhythmic structure repeats at multiple timescales. 
A model that can only attend to recent tokens (or a static adapter at each position) will fail to capture these structural regularities.
Finally, audio processing tasks are generally suitable for SSM models over transformers, a benefit we expect to observe in the HRM adapter.

For this task, we had a backbone frozen TinyGPT (4-layer, d\_model=128), context length T=512 events, with 80K events training, 5K validation.

\begin{figure}[!ht]
    \centering
    \includegraphics[width=1\linewidth]{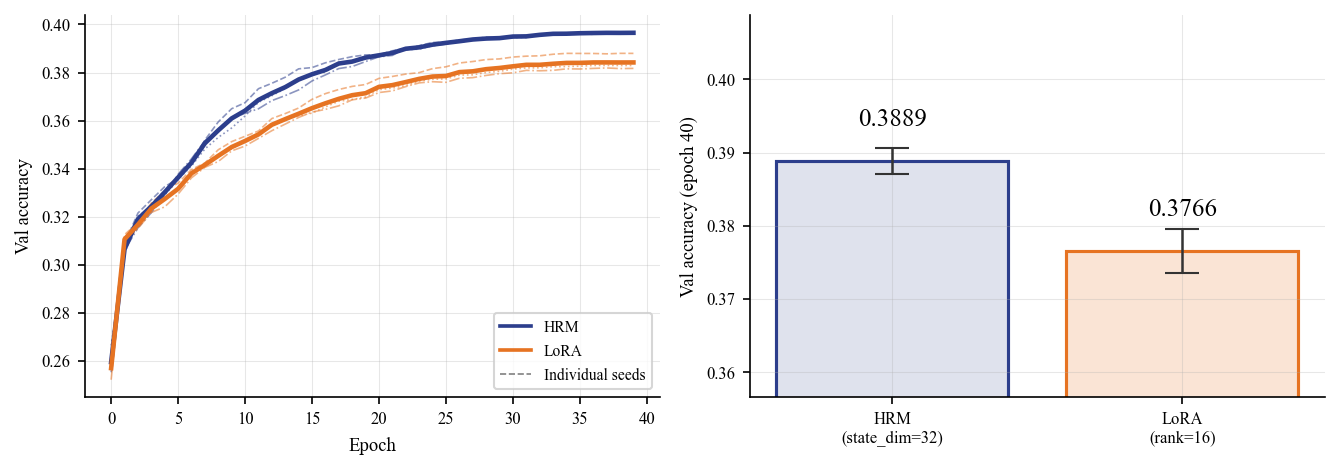}
    \caption{MAESTRO piano language modeling. (left) HRM vs. LoRA, (right) Final BPC. }
    \label{fig:maestro}
\end{figure}

The results are shown in Fig.~\ref{fig:maestro}.
HRM achieves a lower BPC at convergence and with substantially smaller variance across seeds (band is nearly invisible for HRM). (Right) HRM accuracy 0.3966±0.0003 vs LoRA 0.3843±0.0033 at epoch 40, t=7.01, p<0.001. Both adapters use identical parameter budgets (Tier 2, ~33K parameters).

\subsection{\texttt{enwik8} Character Language Modeling}
\texttt{enwik8} is a widely used character level language modeling consisting of 100 million bytes of XML formatted Wikipedia text \cite{mahoney2013large}.
The task is to minimize bits-per-character (BPC), the number of bits required to encode each character on average.
Since English text has word-level, sentence-level, and paragraph-level structure, \texttt{enwik8} is appropriate to test long-range adapter capability for PEFT tasks.
Standard benchmarks use context lengths of 512–8192 characters, which are long by transformer standards. 

For comparing HRM with LoRA, we utilize a backbone frozen TinyGPT (4-layer, $d_{model}=$128), with context lengths $T\in\{512, 1024, 2048\}$, 3 tiers of model capacity ($r$ and $\hat d$ values.
Due to different convergence rates, we use 25 epochs for $T=512$ and 40 epochs for $T=1024$ and $2048$, with a batch size of 32, 10,000 training examples, and 1,000 validation examples.
The findings across these cases are in Table~\ref{tab:bpc-compare} with HRM achieving a lower BPC than LoRA adapter on \emph{all configurations}.
% For low capacity adapter ($r=8,\hat{d}=16$), $\Delta$BPC was found to increase with $T$.
The BPC-$T$ relation is detailed in Appendix~\ref{app:enwik8}.

\begin{remark}\label{remark:tinygpt}
Our model ($\approx$1.1M backbone params + 33K adapter) is not competing with full-scale \texttt{enwik8} models. 
Transformer-XL (44M params) achieves 1.06 BPC; our goal is not SoTA but the relative difference $\Delta$BPC = BPC(LoRA)-BPC(HRM) as $T$ varies. 
A positive growing $\Delta$BPC demonstrates that the HRM can help improve even small capacity TinyGPT backbone on the task.
\end{remark}

\section{Mistral-7B LongBench}
In this final experiment, we evaluate HRM against four different LoRA family baselines: LoRA, AdaLORA, DoRA, and QLoRA.
LongBench is a comprehensive benchmark for evaluating LLMs on their ability to understand and process long-context information across various tasks \cite{bai2024longbench}.
We do this over three different LongBench tasks, using a Mistral-7B-v0.1 pre-trained model with 7.25B parameters \cite{jiang2023mistral}.
The three tasks chosen in LongBench are QuALITY \cite{pang2022quality}, QMSum \cite{zhong2021qmsum}, and NarrativeQA \cite{kovcisky2018narrativeqa}.

QuALITY is a multiple-choice reading comprehension dataset over long articles (avg. $\sim$4,000 tokens), where each example presents an article with a 4-answer multiple choice question.
The fine-tuned model must learn to select the correct option, with the cognitive bottleneck of \emph{sequential evidence integration}.
The model is judged on the top-1 accuracy, i.e., exact match of predicted option letter to gold.

\begin{figure*}[!ht]
  \centering
  \includegraphics[width=0.9\textwidth]{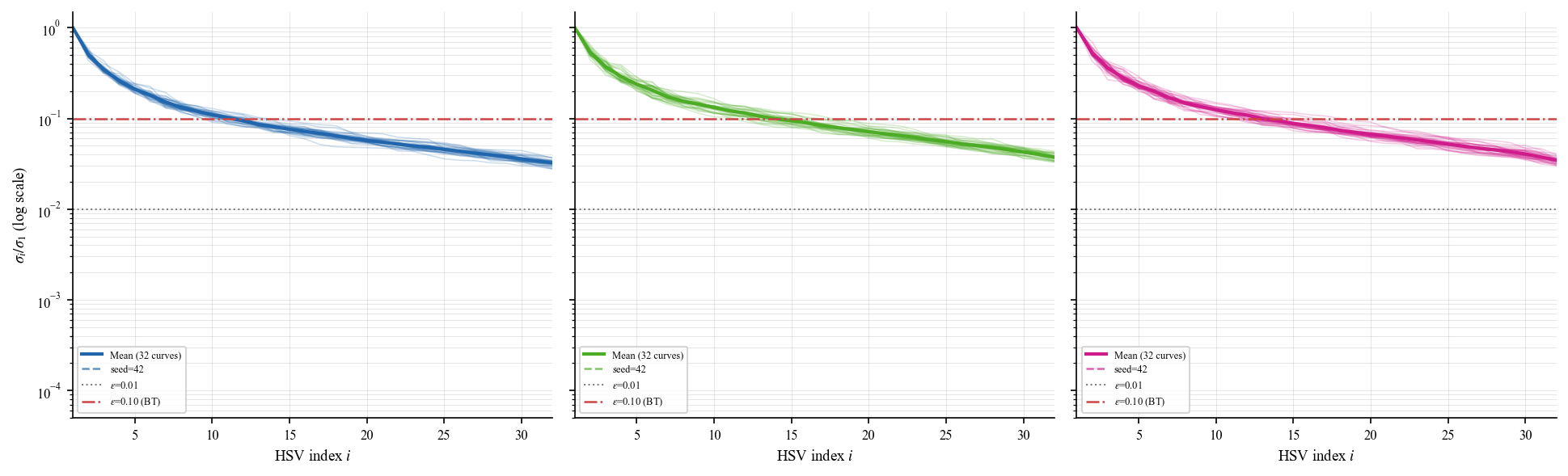} 
  \caption{Mistral-7B HRM: Hankel Singular Value decay curves (left) QuALITY, (middle) QMSum, (right) VarrativeQA.}
  \label{fig:mistral-hsv}
\end{figure*}

QMSum is a query-focused summarization dataset of meeting transcripts (avg. $\sim$10,000 tokens per meeting, truncated to 4096 tokens).
Given a query and a transcript, the fine-tuned model must learn to generate a paragraph-length summary addressing the query.
Since meeting transcripts are inherently sequential, turn-taking, speaker contributions, and topic shifts follow a temporal order that is meaningful for summarization. 
We measure ROUGE-1, ROUGE-2, and ROUGE-L against reference summaries.

Finally, NarrativeQA is an open-ended question answering dataset over full books and movie scripts (avg. $\sim$50,000 tokens per document).
For each document-question pair, the answer is a specific phrase or sentence from the document.
The fine-tuned model must identify and generate the exact answer phrase from within the document context. 
We measure token-level F1 score (case-insensitive, following standard NarrativeQA evaluation).

All methods are iso-parametric, with HRM state dimension $d=32$ and LoRA rank $r=16$, both yielding $\approx 8.4$M trainable parameters, roughly 0.116\% of the total. 
Training protocol was identical across all methods: 5 epochs, lr$=5\times 10^{-4}$, batch size=1, gradient accumulation=8 (effective batch=8), max input length$=4096$, with AdamW optimizer. 
Evaluation uses the held-out 10\% test split for QuALITY and QMSum, and the official test split for NarrativeQA.

We found that of the 3 tasks, HRM outperformed all: LoRA, DoRA, QLoRA, and AdaLoRA on two tasks -- QuALITY and QMSum.
While it drastically underperformed all methods on NarativeQA.
HRM exceeded the best baseline (LoRA \& AdaLoRA) on QuALITY on the accuracy metric by +34.8\% (HRM acc. 47.43\% to 35.18\% for LoRA/AdaLoRA). 
HRM exceeded the best baseline (QLoRA) on QMSum on all 3 metrics: ROUGE-1, ROUGE-2, ROUGE-L.
The ROUGE-1 metric was improved by +54.3\% relative (HRM R-1 0.2531 to 0.1641\% for QLoRA), by +73.59\% relative against DoRA, and by +71.36\% relative for both LoRA and AdaLoRA.
However, it drastically underperformed on NarrativeQA, with an F1 score of 0.0391 against the best model (LoRA, F1=0.1592), under performing by a whole -75.4\% (see Appendix~.\ref{app:longbench} for details).

\begin{remark}
Similar to \ref{remark:tinygpt}, Mistral-7B backbone is not suited for SoTA performance on LongBench, as the goal is to compare adapter improvements across different types of tasks that involve long context reasoning.
The is supported in our findings for the 3 tasks.
\end{remark}

QuALITY requires the model to read a $\sim$4K-token article and answer a multiple-choice question, a prototypical sequential integration problem.
\citet{geva2021transformer} establish that transformer MLP blocks function as key-value memories to perform content integration and associative recall, while attention performs positional retrieval.
Fine-tuning with LoRA modifies the attention's $Q, K, V$ projections, improving the model's ability to attend to relevant spans. 
However, for MCQ reasoning, the pre-trained model can already attend to relevant spans and the bottleneck is integrating evidence across multiple spans into a coherent conclusion. 
The HRM adapter provides a recurrent SSM residual that maintains a running integration of the MLP's content representations across sequence positions.
The relative improvement is therefore consistent with this mechanistic prediction. 

QMSum requires generating a focused summary of a meeting transcript in response to a query.
Due to similar reasoning as above, we can explain HRM's relative gains over the LoRA family for PEFT.

NarrativeQA on the other hand fails completely. 
Part of this was because of the hardware constraints of a dual NVIDIA GeForce RTX 4090 GPUs setup, with a total VRAM of 48GB.
Due to this, the complete NarrativeQA context of 50,000 was not used at all, and truncated to 5,000 for training.
As a result, all methods in comparison failed the benchmark, as the NarrativeQA task was not suitable for the hardware.
As a result, HRM vastly underperformed all LoRA family on the task.

\begin{figure*}[!ht]
  \centering
  \includegraphics[width=0.9\textwidth]{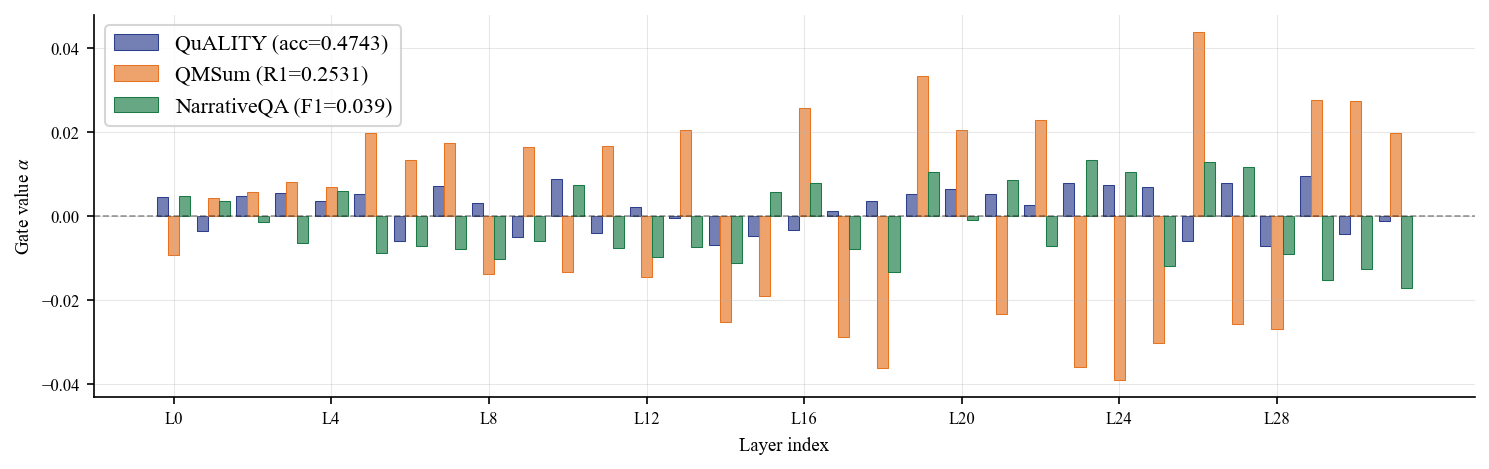} 
  \caption{HRM gate values per layer for Mistral-7B, all three LongBench tasks (gate\_init = 0.1).}
  \label{fig:mistral-gate}
\end{figure*}

At the coarser threshold $\varepsilon=$0.10, the complexity ordering extends to LLM-scale tasks. 
From BT on trained Mistral-7B checkpoints (32 layers), $\hat d$(DFA, $\varepsilon=$0.01)=5 $< \hat d$(QuALITY, $\varepsilon=$0.10) $\approx$11 < $\hat d$(QMSum, $\varepsilon=$0.10)$\approx$13 $<\hat{d}$(\texttt{enwiki8}, $\varepsilon=$0.01)=32 $\leq \hat d$(MAESTRO, $\varepsilon=$0.01)=32.
The HSV decay plot for the three tasks is shown in Fig.~\ref{fig:mistral-hsv}.
Gate analysis reveals that HRM gates converge near zero despite non-zero initialization and explicit weight-decay exclusion: the gradient itself drives closure. 
Yet frozen-gate probing confirms that the trained SSM weights $(A, B, C)$ retain 1180$\times$ larger contribution capacity at gate=0.1
This is shown in the final gate values learned for each layer, for each of the three tasks in Fig.~\ref{fig:mistral-gate}.

\section{Conclusion and Discussion}
In this work, we proposed Hankel reduced order model (HRM) adapter for parameter efficient fine-tuning (PEFT). 
HRM adds a provably compressible recurrent temporal state to any frozen pre-trained transformer backbone, and unlike prior PEFT methods, the adapter is temporal causal.
The model reduction is achieved via balanced truncation of the underlying linear system, where we utilize controllability and observability Grammians from control theory.
This allows us to have a (tight) error bound making model reduction a certified compression, and not a heuristic.
PEFT literature has inadvertently restricted itself to the class of zero-memory adapters.
HRM results show that a gated SSM, inserted parallel to a frozen transformer's MLP blocks, can be trained efficiently, compressed with theoretical guarantees, and consistently outperforms the best static alternative on temporal tasks.
We demonstrate that HRM outperforms LoRA, QLoRA, AdaLoRA, and DoRA on 6 different tasks, from three qualitatively different task families that share the requirement of causal state accumulation
We also observe as a by-product that Hankel singular values and associated Grammians are a strong metric for the training task's memory requirements.
% \paragraph{Limitations Observed}

\paragraph{Open Problems \& Next Steps}
Numerous immediate next steps emerge from the presented HRM works.
1. Extension of Hankel singular value-based balanced truncation to selective SSMs (Mamba-type, with input-dependent $\bar{A}_t)$ benefits the proposed adapter.
This would involve computationally tractable parameter-varying empirical Grammian computations that do not cause overhead larger than the intermediate FFT/IFFT calculations.
2. Currently the BT compression takes place after an initial phase's SSM adapter training.
An enhancement would be to \emph{adaptively allocate Hankel ranks for each layer}, i.e., $\hat{d}^{(l)}$ for layer $l$.
This is analogous to AdaLoRA's rank allocation, but would be informed by per-layer $\sigma_i^{(l)}$ HSV spectra.
3. More complete benchmarking on LongBench of existing results is needed, for more tasks, across deeper context lengths.
4. Mixed injection of the SSM adapter needs to be ablated: attention injection site vs. MLP injection site, and more importantly, finding out task signatures suitable for each injection site for the adapter, and examine if simultaneous injection would benefit retrieval + integration tasks.
5. Investigate frozen gate training scenarios (e.g., \texttt{gate.requires\_grad=False}) to enforce sustained HRM adapter contribution, and directly test whether larger training times translate to performance gains.

\section*{Impact Statement}
This paper presents work whose goal is to advance the field of Machine
Learning. There are many potential societal consequences of our work, none
which we feel must be specifically highlighted here.


\bibliography{bibliography}

@article{hu2022lora,
  title={Lora: Low-rank adaptation of large language models.},
  author={Hu, Edward J and Shen, Yelong and Wallis, Phillip and Allen-Zhu, Zeyuan and Li, Yuanzhi and Wang, Shean and Wang, Liang and Chen, Weizhu and others},
  journal={Iclr},
  volume={1},
  number={2},
  pages={3},
  year={2022}
}

@inproceedings{liu2024dora,
  title={Dora: Weight-decomposed low-rank adaptation},
  author={Liu, Shih-Yang and Wang, Chien-Yi and Yin, Hongxu and Molchanov, Pavlo and Wang, Yu-Chiang Frank and Cheng, Kwang-Ting and Chen, Min-Hung},
  booktitle={Forty-first International Conference on Machine Learning},
  year={2024}
}

@article{hawthorne2018enabling,
  title={Enabling factorized piano music modeling and generation with the MAESTRO dataset},
  author={Hawthorne, Curtis and Stasyuk, Andriy and Roberts, Adam and Simon, Ian and Huang, Cheng-Zhi Anna and Dieleman, Sander and Elsen, Erich and Engel, Jesse and Eck, Douglas},
  journal={arXiv preprint arXiv:1810.12247},
  year={2018}
}

@article{zhang2023adalora,
  title={Adalora: Adaptive budget allocation for parameter-efficient fine-tuning},
  author={Zhang, Qingru and Chen, Minshuo and Bukharin, Alexander and Karampatziakis, Nikos and He, Pengcheng and Cheng, Yu and Chen, Weizhu and Zhao, Tuo},
  journal={arXiv preprint arXiv:2303.10512},
  year={2023}
}

@article{dettmers2023qlora,
  title={Qlora: Efficient finetuning of quantized llms},
  author={Dettmers, Tim and Pagnoni, Artidoro and Holtzman, Ari and Zettlemoyer, Luke},
  journal={Advances in neural information processing systems},
  volume={36},
  pages={10088--10115},
  year={2023}
}

@article{hayou2024lora+,
  title={Lora+: Efficient low rank adaptation of large models},
  author={Hayou, Soufiane and Ghosh, Nikhil and Yu, Bin},
  journal={arXiv preprint arXiv:2402.12354},
  year={2024}
}

@article{liu2022few,
  title={Few-shot parameter-efficient fine-tuning is better and cheaper than in-context learning},
  author={Liu, Haokun and Tam, Derek and Muqeeth, Mohammed and Mohta, Jay and Huang, Tenghao and Bansal, Mohit and Raffel, Colin A},
  journal={Advances in Neural Information Processing Systems},
  volume={35},
  pages={1950--1965},
  year={2022}
}

@inproceedings{lester2021power,
  title={The power of scale for parameter-efficient prompt tuning},
  author={Lester, Brian and Al-Rfou, Rami and Constant, Noah},
  booktitle={Proceedings of the 2021 conference on empirical methods in natural language processing},
  pages={3045--3059},
  year={2021}
}

@inproceedings{li2021prefix,
  title={Prefix-tuning: Optimizing continuous prompts for generation},
  author={Li, Xiang Lisa and Liang, Percy},
  booktitle={Proceedings of the 59th Annual Meeting of the Association for Computational Linguistics and the 11th International Joint Conference on Natural Language Processing (Volume 1: Long Papers)},
  pages={4582--4597},
  year={2021}
}

@inproceedings{houlsby2019parameter,
  title={Parameter-efficient transfer learning for NLP},
  author={Houlsby, Neil and Giurgiu, Andrei and Jastrzebski, Stanislaw and Morrone, Bruna and De Laroussilhe, Quentin and Gesmundo, Andrea and Attariyan, Mona and Gelly, Sylvain},
  booktitle={International conference on machine learning},
  pages={2790--2799},
  year={2019},
  organization={PMLR}
}

@inproceedings{pfeiffer2020adapterhub,
  title={Adapterhub: A framework for adapting transformers},
  author={Pfeiffer, Jonas and R{\"u}ckl{\'e}, Andreas and Poth, Clifton and Kamath, Aishwarya and Vuli{\'c}, Ivan and Ruder, Sebastian and Cho, Kyunghyun and Gurevych, Iryna},
  booktitle={Proceedings of the 2020 conference on empirical methods in natural language processing: system demonstrations},
  pages={46--54},
  year={2020}
}

@inproceedings{gu2024mamba,
  title={Mamba: Linear-time sequence modeling with selective state spaces},
  author={Gu, Albert and Dao, Tri},
  booktitle={First conference on language modeling},
  year={2024}
}

@article{gu2022parameterization,
  title={On the parameterization and initialization of diagonal state space models},
  author={Gu, Albert and Goel, Karan and Gupta, Ankit and R{\'e}, Christopher},
  journal={Advances in neural information processing systems},
  volume={35},
  pages={35971--35983},
  year={2022}
}

@article{gu2021efficiently,
  title={Efficiently modeling long sequences with structured state spaces},
  author={Gu, Albert and Goel, Karan and R{\'e}, Christopher},
  journal={arXiv preprint arXiv:2111.00396},
  year={2021}
}

@article{gu2023mamba,
  title={Mamba: Linear-time sequence modeling with selective state spaces},
  author={Gu, Albert and Dao, Tri},
  journal={arXiv preprint arXiv:2312.00752},
  year={2023}
}

@article{de2024griffin,
  title={Griffin: Mixing gated linear recurrences with local attention for efficient language models},
  author={De, Soham and Smith, Samuel L and Fernando, Anushan and Botev, Aleksandar and Cristian-Muraru, George and Gu, Albert and Haroun, Ruba and Berrada, Leonard and Chen, Yutian and Srinivasan, Srivatsan and others},
  journal={arXiv preprint arXiv:2402.19427},
  year={2024}
}

@article{park2024can,
  title={Can mamba learn how to learn? a comparative study on in-context learning tasks},
  author={Park, Jongho and Park, Jaeseung and Xiong, Zheyang and Lee, Nayoung and Cho, Jaewoong and Oymak, Samet and Lee, Kangwook and Papailiopoulos, Dimitris},
  journal={arXiv preprint arXiv:2402.04248},
  year={2024}
}

@article{lieber2024jamba,
  title={Jamba: A hybrid transformer-mamba language model},
  author={Lieber, Opher and Lenz, Barak and Bata, Hofit and Cohen, Gal and Osin, Jhonathan and Dalmedigos, Itay and Safahi, Erez and Meirom, Shaked and Belinkov, Yonatan and Shalev-Shwartz, Shai and others},
  journal={arXiv preprint arXiv:2403.19887},
  year={2024}
}

@inproceedings{pang2022quality,
  title={QuALITY: Question answering with long input texts, yes!},
  author={Pang, Richard Yuanzhe and Parrish, Alicia and Joshi, Nitish and Nangia, Nikita and Phang, Jason and Chen, Angelica and Padmakumar, Vishakh and Ma, Johnny and Thompson, Jana and He, He and others},
  booktitle={Proceedings of the 2022 Conference of the North American Chapter of the Association for Computational Linguistics: Human Language Technologies},
  pages={5336--5358},
  year={2022}
}

@inproceedings{geva2021transformer,
  title={Transformer feed-forward layers are key-value memories},
  author={Geva, Mor and Schuster, Roei and Berant, Jonathan and Levy, Omer},
  booktitle={Proceedings of the 2021 Conference on Empirical Methods in Natural Language Processing},
  pages={5484--5495},
  year={2021}
}

@inproceedings{zhong2021qmsum,
  title={QMSum: A new benchmark for query-based multi-domain meeting summarization},
  author={Zhong, Ming and Yin, Da and Yu, Tao and Zaidi, Ahmad and Mutuma, Mutethia and Jha, Rahul and Hassan, Ahmed and Celikyilmaz, Asli and Liu, Yang and Qiu, Xipeng and others},
  booktitle={Proceedings of the 2021 Conference of the North American Chapter of the Association for Computational Linguistics: Human Language Technologies},
  pages={5905--5921},
  year={2021}
}

@article{kovcisky2018narrativeqa,
  title={The narrativeqa reading comprehension challenge},
  author={Ko{\v{c}}isk{\`y}, Tom{\'a}{\v{s}} and Schwarz, Jonathan and Blunsom, Phil and Dyer, Chris and Hermann, Karl Moritz and Melis, G{\'a}bor and Grefenstette, Edward},
  journal={Transactions of the Association for Computational Linguistics},
  volume={6},
  pages={317--328},
  year={2018},
  publisher={MIT Press One Rogers Street, Cambridge, MA 02142-1209, USA journals-info~…}
}

@article{jiang2023mistral,
  title={Mistral 7B},
  author={Jiang, Albert Q and Sablayrolles, Alexandre and Mensch, Arthur and Bamford, Chris and Chaplot, Devendra Singh and de las Casas, Diego and Bressand, Florian and Lengyel, Gianna and Lample, Guillaume and Saulnier, Lucile and others},
  journal={arXiv preprint arXiv:2310.06825},
  year={2023},
  url={https://arxiv.org/abs/2310.06825}
}

@article{sheng2024slora,
  title={Slora: Scalable serving of thousands of lora adapters},
  author={Sheng, Ying and Cao, Shiyi and Li, Dacheng and Hooper, Coleman and Lee, Nicholas and Yang, Shuo and Chou, Christopher and Zhu, Banghua and Zheng, Lianmin and Keutzer, Kurt and others},
  journal={Proceedings of Machine Learning and Systems},
  volume={6},
  pages={296--311},
  year={2024}
}

@inproceedings{bai2024longbench,
  title={Longbench: A bilingual, multitask benchmark for long context understanding},
  author={Bai, Yushi and Lv, Xin and Zhang, Jiajie and Lyu, Hongchang and Tang, Jiankai and Huang, Zhidian and Du, Zhengxiao and Liu, Xiao and Zeng, Aohan and Hou, Lei and others},
  booktitle={Proceedings of the 62nd annual meeting of the association for computational linguistics (volume 1: Long papers)},
  pages={3119--3137},
  year={2024}
}

@article{mahoney2013large,
  title={Large text compression benchmark, 2011},
  author={Mahoney, Matt},
  journal={URL http://www. mattmahoney. net/dc/text. html},
  year={2013}
}

@article{zhang2023loraprune,
  title={Loraprune: Pruning meets low-rank parameter-efficient fine-tuning},
  author={Zhang, Mingyang and Chen, Hao and Shen, Chunhua and Yang, Zhen and Ou, Linlin and Yu, Xinyi and Zhuang, Bohan},
  year={2023}
}

@article{hao2024flora,
  title={Flora: Low-rank adapters are secretly gradient compressors},
  author={Hao, Yongchang and Cao, Yanshuai and Mou, Lili},
  journal={arXiv preprint arXiv:2402.03293},
  year={2024}
}

@article{lall1999empirical,
  title={Empirical model reduction of controlled nonlinear systems},
  author={Lall, Sanjay and Marsden, Jerrold E and Glava{\v{s}}ki, Sonja},
  journal={IFAC Proceedings Volumes},
  volume={32},
  number={2},
  pages={2598--2603},
  year={1999},
  publisher={Elsevier}
}

@book{corless2003linear,
  title={Linear systems and control: an operator perspective},
  author={Corless, Martin J and Frazho, Art},
  year={2003},
  publisher={CRC Press}
}

@article{glover1984all,
  title={All optimal Hankel-norm approximations of linear multivariable systems and their {$L^\infty$}-error bounds},
  author={Glover, Keith},
  journal={International journal of control},
  volume={39},
  number={6},
  pages={1115--1193},
  year={1984},
  publisher={Taylor \& Francis}
}

@article{moore2003principal,
  title={Principal component analysis in linear systems: Controllability, observability, and model reduction},
  author={Moore, Bruce},
  journal={IEEE transactions on automatic control},
  volume={26},
  number={1},
  pages={17--32},
  year={2003},
  publisher={IEEE}
}

@inproceedings{aghajanyan2021intrinsic,
  title={Intrinsic dimensionality explains the effectiveness of language model fine-tuning},
  author={Aghajanyan, Armen and Gupta, Sonal and Zettlemoyer, Luke},
  booktitle={Proceedings of the 59th annual meeting of the association for computational linguistics and the 11th international joint conference on natural language processing (volume 1: long papers)},
  pages={7319--7328},
  year={2021}
}
\bibliographystyle{icml2026}

%%%%%%%%%%%%%%%%%%%%%%%%%%%%%%%%%%%%%%%%%%%%%%%%%%%%%%%%%%%%%%%%%%%%%%%%%%%%%%%

% APPENDIX

%%%%%%%%%%%%%%%%%%%%%%%%%%%%%%%%%%%%%%%%%%%%%%%%%%%%%%%%%%%%%%%%%%%%%%%%%%%%%%%
\newpage
\appendix
\onecolumn
\section{Iso-parameter Comparison}\label{app:iso-param}
In order to compare HRM and LoRA on an equal footing, we ensure that $r$ and $\hat{d}$ are chosen such that $\abs{P_{LoRA} - P_{HRM}}\leq 0.1\%$.
Such an iso-parametric table to choose $r$ and $\hat{d}$ is shown below.
All experiments in the paper use all three tiers to demonstrate consistency, and conclusions are drawn only from the pattern across tiers rather than individual tier performance.

\begin{table}[htbp]
\centering
\caption{Comparison of parameter counts between LoRA and HRM configurations.}
\label{tab:parameter_comparison}
\renewcommand{\arraystretch}{1.2} % Improves row spacing for better readability
\begin{tabular}{rrcrcc}
\toprule
\textbf{LoRA $r$} & \textbf{$P_{\text{LoRA}}$} & \textbf{HRM $\hat{d}$} & \textbf{$P_{\text{HRM}}$} & \textbf{$\Delta$ (\%)} & Regime\\
\midrule
8  & 16,384 & 16 & 16,156 & $+0.81$ & Low capacity \\
16 & 32,768 & 32 & 33,028 & $+0.79$ & Mid capacity\\
32 & 65,536 & 63 & 65,020 & $-0.79$ & High capacity \\
\bottomrule
\end{tabular}
\end{table}

\section{Parallel Scan for HRM}\label{app:fft-compute-time}
Since $\bar{A}$ is diagonal with entries $\bar{a}_1,\cdots,\bar{a}_d$, the impulse response factorizes: $g_k = C \bar{A}^k \bar{B} = C \mathrm{diag}(\bar{a}_1^d,\cdots,\bar{a}_d^d)\bar{B}$. 
The $i^{\mathrm{th}}$ column of $\bar{A}^k\bar{B}=\bar{a}_i^kcdot \bar{B}_i$, so the full convolution can be implemented as $d$ scalar convolutions (one per state dimension), each with a geometric impulse response. 
This is the efficient form used in the code. 
On the other hand, the FFT forward pass computes for each state dimension $i$, $g^{(i)}_k=\bar{a}_i^k$ for $k=0,\cdots, T-1$, being $\bigO{T}$ per dimension.
Next, FFT($\tilde g^{(i)}$) and $\tilde h^{(i)}$ is $\bigO{T\log T}$ per dimension, giving a total of $\bigO{T \log T}$.
The element-wise product $\odot$ and IFFT computations in (\ref{eq:fft-output}) are $\bigO{d T\log T}$, and the final $\bar{B},C$ projections are $\bigO{d\cdot d_{model}\cdot T}$.
The net complexity for the FFT forward pass is therefore, $\bigO{d_{model}dT + dT\log T}$.
For example, if the model order $d$, or $\hat{d}$ is 16, then $d\ll d_{model}=128$, the next complexity if dominated by $\bigO{d_{model} \cdot d\cdot T}$.

\begin{table}[!htbp]
\centering
\caption{Epoch training times on MacBook Pro M2 (MPS backend). HRM-sequential is 10–14$\times$ slower than LoRA. HRM-FFT matches LoRA exactly at tested context lengths.}
\label{tab:fft-compute-compare}
\renewcommand{\arraystretch}{1.3} % Slightly more breathing room for wrapped headers

\begin{tabularx}{\textwidth}{>{\raggedleft\arraybackslash}p{1.5cm} X X c X c}
\toprule
\textbf{Context Length $T$} & \textbf{LoRA Wall clock time/epoch (s)} & \textbf{HRM-seq. Wall clock time/epoch} & \textbf{HRM-seq. / LoRA} & \textbf{HRM-FFT Wall clock time/epoch} & \textbf{HRM-FFT / LoRA} \\
\midrule
512  & 9   & $\sim$90   & $10\times$ & 9   & $1\times$ \\
1024 & 70  & $\sim$980  & $14\times$ & 70  & $1\times$ \\
2048 & 265 & $\sim$3710 & $14\times$ & 265 & $1\times$ \\
\bottomrule
\end{tabularx}
\end{table}

\begin{wrapfigure}{!r}{0.4\textwidth} % {alignment}{width}
  \centering
  \includegraphics[width=0.38\textwidth]{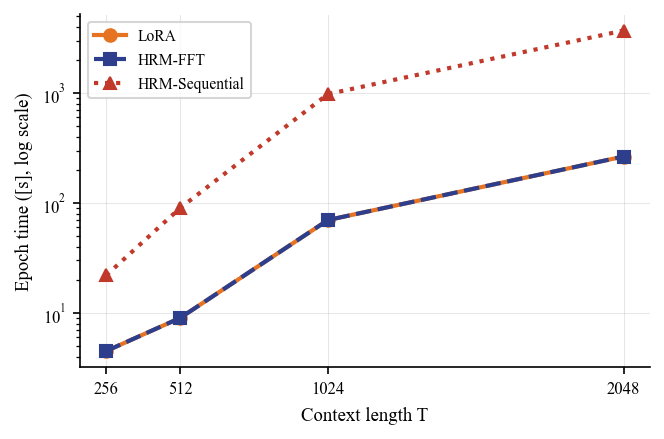}
  \caption{Maximum absolute error between FFT and sequential scan outputs over 100 random input sequences at varying T.}
  \label{fig:fft-exactness}
\end{wrapfigure}
Since FFT operations in \texttt{float32} introduce rounding errors of order $10^{-7}$. We verify empirically compare sequential and FFT outputs across 100 random sequences.
We find that the maximum absolute error is $<5\times 10^{-6}$, negligible for gradient computation, confirming the FFT equivalence is exact up to floating-point rounding (Fig.~\ref{fig:fft-exactness}).

We compared the compute for with and without the FFT-based parallel scan over varying context lengths.
Training at T=2048 with batch=32 required issuing 2048 Python-level dispatcher calls per layer per forward pass. At 4 layers and batch=32, this is 4 × 2048 × 32 = 262,144 sequential operations — completely saturating the Python-PyTorch overhead.

\section{Is there a Case for SVD over HSVs?}\label{app:why-hsv}
A singular value decomposition to reduce model order seems like a natural alternative to balanced truncation.
Several works apply this idea to LoRA matrices: FLORA \cite{hao2024flora}, LoRA pruning \cite{zhang2023loraprune}. 
We argue that SVD is the wrong tool for a dynamical system, and BT is the correct one.
For static matrix order/rank reduction, SVD is a more natural choice.
For instance, a state direction may have a large singular value in $\bar{B}$ (i.e., the input strongly drives that direction) yet be completely unobservable, i.e., producing zero output at channel corresponding to $C$.
SVD of $\bar{B}$ would retain this direction as important, whereas its Hankel singular value (HSV) would be $\sigma_i=0$, thereby getting discarded.
Conversely, a direction with small $\bar{B}$ singular value might be the only one observable at $C$, and SVD would drop it while its HSV would preserves it.

\begin{wrapfigure}{!r}{0.45\textwidth} % {alignment}{width}
  \centering
  \includegraphics[width=0.45\textwidth]{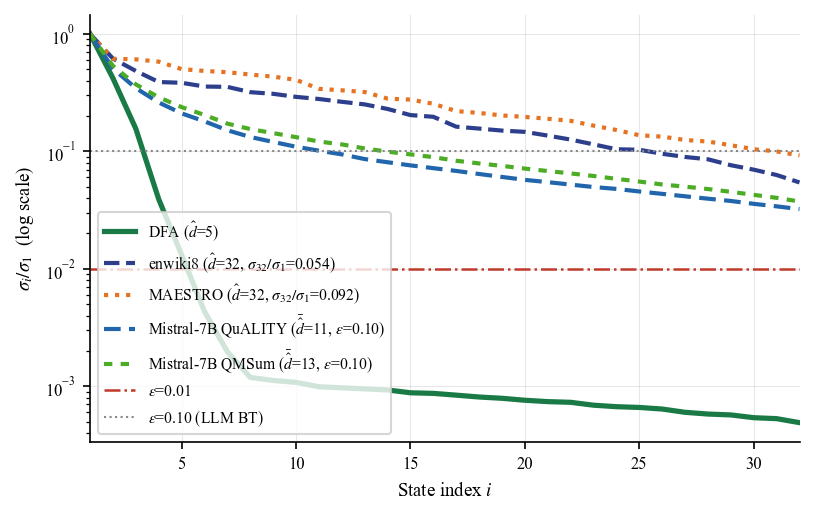}
  \caption{HSV decay for: DFA, \texttt{enwik8}, MAESTRO, QuALITY, and QMSum.}
  \label{fig:hsv-decay-all}
\end{wrapfigure}

SVD identifies directions that are large in individual matrices. HSVs identify directions that are simultaneously reachable and observable in the complete input-output system. 
Due to the causal input-output relation encoded by the Hankel operator, a singular value of $\bar{B}$ alone would not be able to drive model reduction.
Only the Grammian product $W_cW_o$ gives the correct importance measure in this case, since accounts for the full dynamics.
We empirically confirm this on the DFA task, where the largest singular value of $\bar{B}$ does not correspond to the dominant Hankel singular value $\sigma_1$. 
If we had truncated by SVD($\bar{B}$), we would have retained different state directions than BT and achieved a worse compression ratio. 
The BT result $\hat{d}$=6 with $<0.3\%$ accuracy loss would likely not have been achievable by SVD of any single weight matrix.

\subsection{Further Insights from HSVs}\label{app:hsv-insight}
Since the Hankel operator encodes the dynamical input-output relation as $\mathcal H:\text{\{past inputs\}} \to \text{\{future outputs\}}$, the HSV decay curve $\sigma_i/\sigma_1$ characterizes the task's temporal complexity.
For instance, a steep decay curve signifies that most of the temporal complexity of the dynamics are contained in very few dynamical modes, therefore, $d\to\hat d$ HRM is highly compressible.
Conversely, a gradual decay indicates a genuine high-dimensional memory requirement.
To this end, \emph{HSVs, and associated (empirical) Grammians are a strong metric for the training task's memory requirement}.
Fig.~\ref{fig:hsv-decay-all} collects all available HSV decay results presented in this work. 

\section{\texttt{enwik8} Task}\label{app:enwik8}

\begin{figure*}[!htbp]
  \centering
  \includegraphics[width=0.875\textwidth]{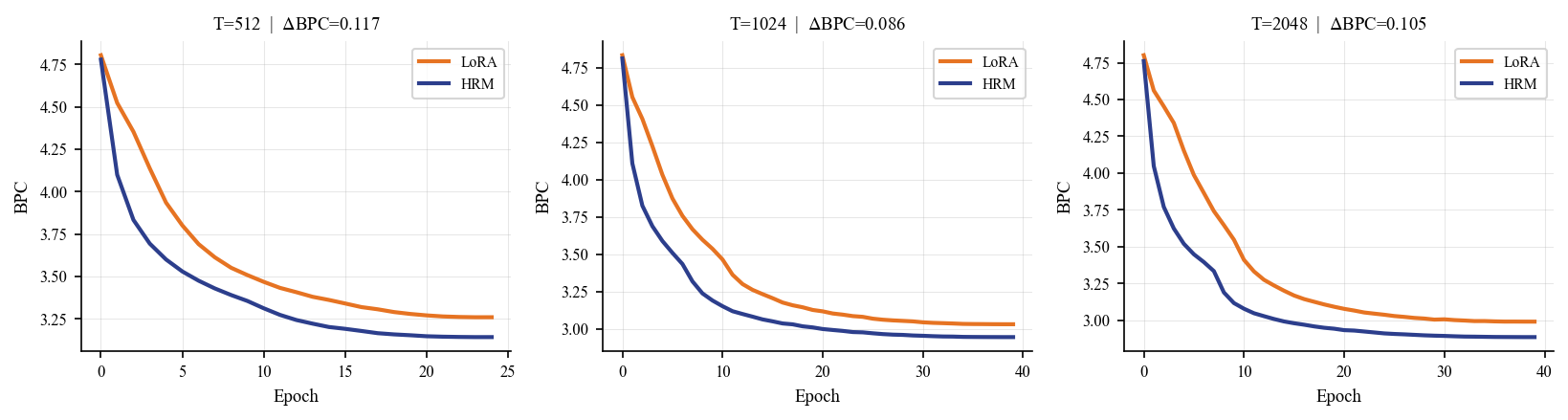} 
  \caption{\texttt{enwiki8} BPC learning curves at Tier 2 (HRM $d=32$ vs LoRA $r=16$) for T $\in$ {512, 1024, 2048}. The region between curves represents the BPC advantage of HRM over LoRA.}
  \label{fig:enwik8}
\end{figure*}

The HRM adapter's dominant state mode has a learned eigenvalue $\bar{a}_{\max}$ ($\bar{a}_{\max} \approx$0.97–0.99 after training). 
The fraction of signal retained from a token k steps ago is $\bar{a}_{\max}^k$. 
At T=512, the adapter retains $\bar{a}_{\max}^{256}\approx 0.97^{256} \approx 0.0006$ of signal from the midpoint of the context window.
This means that only the most recent $\sim$70 tokens contribute substantially (at 1\% threshold). 
At $T=$2048, the same adapter covers proportionally more of the window — now $\bar{a}_{\max}^{1024} \approx 10^{-13}$, but $\bar{a}_{\max}^70\approx$0.11 still holds. 
The key insight is that the adapter's effective memory depth is fixed by the trained eigenvalues, but the fraction of context this covers grows with $T$. 
At $T$=2048, the 70-step memory covers 3.4\% of context; at $T$=512, it covers 13.7\%. 
The LoRA static $\Delta W$ covers none. 
As $T$ grows, HRM's absolute reach stays roughly constant while LoRA's relative disadvantage grows.

\begin{table}[htbp]
\centering
\caption{Comparison of BPC across different adapter capacities and context lengths.}
\label{tab:bpc-compare}
\renewcommand{\arraystretch}{1.2} 
\begin{tabular}{l|c|c|c|c}
\hline
\textbf{Adapter Capacity} & \textbf{Context length} & \textbf{LoRA BPC} & \textbf{HRM BPC} & \textbf{$\Delta$BPC} \\ \hline
% We use \raisebox{-height}{text} to lower the text into the middle row
\raisebox{-2.5ex}{$r=18,\hat{d}=16$} & 512  & 3.3006 & 3.1902 & 0.1104 \\ \cline{2-5} 
                                     & 1024 & 3.0271 & 2.9831 & 0.0440 \\ \cline{2-5} 
                                     & 2048 & 3.0039 & 2.9026 & 0.1013 \\ \hline
\raisebox{-2.5ex}{$r=16,\hat{d}=32$} & 512  & 3.2594 & 3.1424 & 0.1170 \\ \cline{2-5} 
                                     & 1024 & 3.0326 & 2.9464 & 0.0862 \\ \cline{2-5} 
                                     & 2048 & 2.9911 & 2.8859 & 0.1052 \\ \hline
\raisebox{-2.5ex}{$r=32,\hat{d}=63$} & 512  & 3.1789 & 3.1470 & 0.0319 \\ \cline{2-5} 
                                     & 1024 & 3.0186 & 2.9413 & 0.0773 \\ \cline{2-5} 
                                     & 2048 & 2.9906 & 2.8789 & 0.1117 \\ \hline
\end{tabular}
\end{table}

\section{Parity Task}
The parity task requires predicting, at each position $t$, the parity bit 
$p_t=\oplus_{k\leq t} \sigma_k (\text{mod }2)$ where $\sigma_k\in\{0,1\}$. Unlike DFA state tracking (which requires tracking $k=4$ states), parity has a minimal state of exactly 1 bit. 
An ideal adapter would learn $\hat d$=1. 
In practice, HRM converges to $\hat d \approx$ 6-7 (BT threshold $\varepsilon$=0.01), which is higher than expected, suggesting the model learns redundant but numerically stable representations of the parity state. 
The task serves as a lower bound: if HRM cannot outperform LoRA here, it is unlikely to help on any sequential task.

\begin{figure}[!htbp]
\centering
\includegraphics[width=0.75\linewidth]{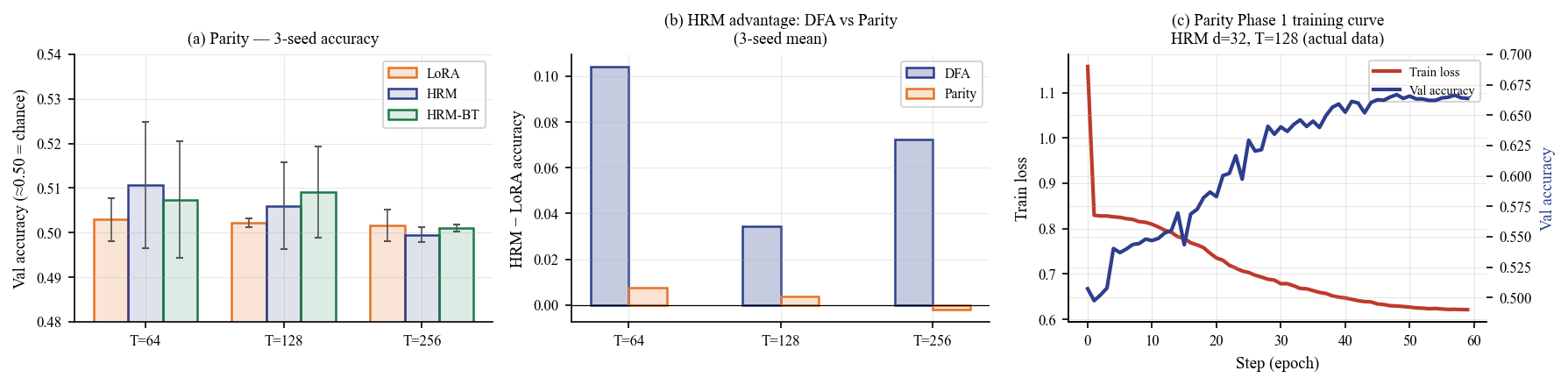}
\caption{(left) Parity accuracy at medium model capacity, 3-seed mean $\pm$ std. All adapters near chance (0.50), parity is near-intractable for a small frozen backbone at $T=256$. (middle) HRM advantage (HRM mean-LoRA mean) on DFA vs. parity, (right) training curve}.
\label{fig:parity}
\end{figure}
Our observations (shown in Fig.~\ref{fig:parity}) show that DFA exhibits a large, advantage with T while parity shows essentially zero HRM benefit. 
This contrast validates the memory hypothesis: HRM helps when multi-dimensional state is required, not when the task can be solved by single-bit counting.

\section{LongBench Tasks}\label{app:longbench}

\begin{table}[htbp]
\centering
\caption{Comparison of HRM against baselines on LongBench: QuALITY, QMSum, NarrativeQA}
\label{tab:hrm-longbench}
\renewcommand{\arraystretch}{1.3}
\begin{tabular}{l|c|l|c}
\hline
\textbf{Method} & \textbf{QuALITY (Acc)} & \textbf{QMSum (R1/R2/R)} & \textbf{NarrativeQA (F1)} \\ \hline
LoRA    & 0.3518 & 0.1477 / 0.0133 / 0.0925 & 0.1592 \\
AdaLoRA & 0.3518 & 0.1477 / 0.0133 / 0.0925 & 0.1571 \\ 
DoRA    & 0.3478 & 0.1458 / 0.0133 / 0.0916 & 0.1569 \\ 
QLoRA   & 0.3399 & 0.1641 / 0.0140 / 0.1038 & 0.1492 \\ 
HRM     & 0.4743 & \textbf{0.2531 / 0.0339 / 0.1180} & 0.1492 \\ \hline
% Using a small font or manual wrap for the comparison row if it's too wide
\textit{HRM vs. best baseline} & \textit{+0.1225 (+34.8\%)} & \textit{+0.0890 R-1 (+54.3\% vs LoRA)} & \textit{-0.1201 (-75.4\%)} \\ \hline
\end{tabular}
\end{table}

\section{Ablations}
\paragraph{BT Threshold $\varepsilon$}
The BT threshold $\varepsilon$ controls the trade-off between compression ratio ($\hat{d}/d$) and accuracy loss. A smaller $\varepsilon$ means fewer dimensions pruned, and a higher $\hat{d}$ implies a better accuracy but less compression. 
Conversely, a larger $\varepsilon$ results in more aggressive pruning.
We capture the ablation in the Figure below.
Fig.~\ref{fig:dfa-ablate} shows that $\hat d$ as a function of $\hat d$ for DFA at T $\in$ {64,128,256,512}: $\hat d$ drops from 30–32 at $\hat d$=0.001 to 2–5 at $\hat d$=0.2.
The elbow in the $\hat d$ vs $\hat d$ curve near $\hat d$=0.01 is the natural compression point, further pruning causes accuracy degradation (panel b). 
Default $\hat d$=0.01 is chosen at this elbow.
\vspace{-0.7cm}
\begin{figure}[h]
    \centering
    \begin{subfigure}[b]{0.325\textwidth}
        \centering
        \includegraphics[width=\textwidth]{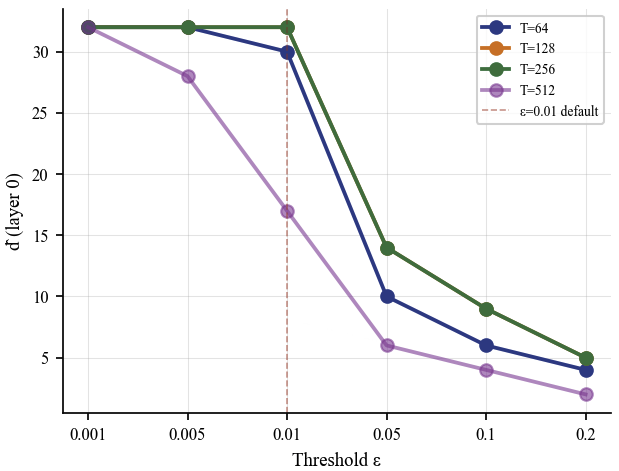}
        % \caption{Left}
    \end{subfigure}
    % \hfill
    \begin{subfigure}[b]{0.325\textwidth}
        \centering
        \includegraphics[width=\textwidth]{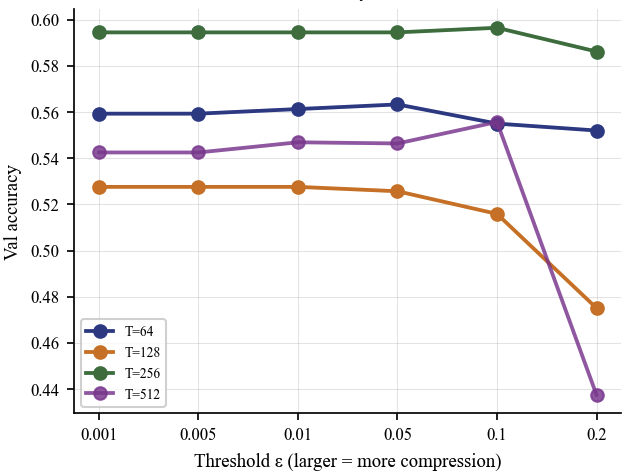}
        % \caption{Right}
    \end{subfigure}
    \caption{BT threshold ablation on DFA (left) DFA: $\hat d$ vs $\hat d$ threshold (layer 0), (right) DFA: accuracy vs $\varepsilon$ threshold.}\label{fig:dfa-ablate}
\end{figure}

\begin{wrapfigure}{!r}{0.35\textwidth} % {alignment}{width}
  \centering
  \includegraphics[width=0.35\textwidth]{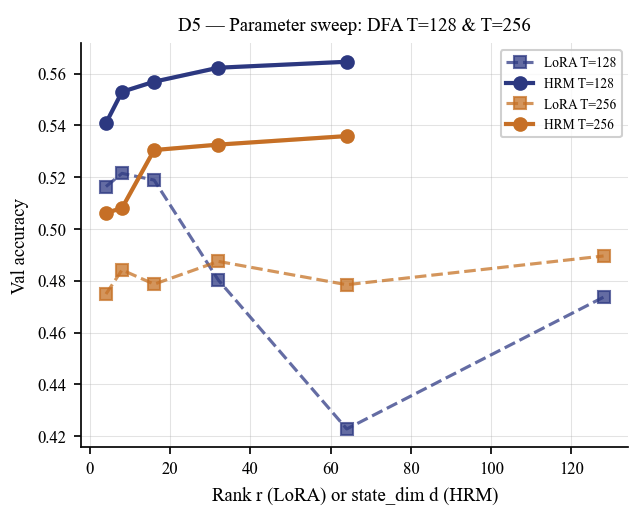}
  \caption{DFA T=128 and T=256: LoRA (rank) vs HRM (state\_dim).}
  \label{fig:rank-ablate}
\end{wrapfigure}
\paragraph{Rank vs. Accuracy}
To verify that HRM's advantage is not merely a parameter count artifact, we sweep LoRA rank $r\in$ \{4,8,16,32,64,128\} and HRM state\_dim d $\in$ \{4,8,16,32,64\} on DFA at T=128 and T=256.
From Fig.~\ref{fig:rank-ablate}, HRM (blue) shows monotonically improving accuracy with state\_dim on DFA. 
LoRA (orange) plateaus and even degrades at high rank ($r=$64,128), a sign of overfitting to position-level features. 
The HRM accuracy at $d=$32 (0.562) exceeds the best LoRA at any rank (0.522 at $r=$16), confirming the advantage is structural not parametric.

\begin{wrapfigure}{!r}{0.35\textwidth} % {alignment}{width}
  \centering
  \includegraphics[width=0.35\textwidth]{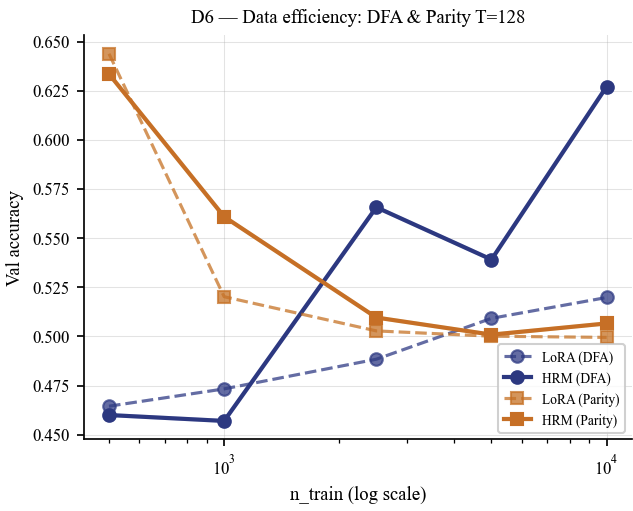}
  \caption{Data efficiency for DFA and Parity T=128: val accuracy vs. n\_train.}
  \label{fig:data-efficiency-ablate}
\end{wrapfigure}

\paragraph{Data Efficiency}
HRM requires training to discover useful dynamics, followed by balanced truncation. 
We test whether HRM is less data-efficient than LoRA at small n\_train.
Fig.~\ref{fig:data-efficiency-ablate} shows that on DFA task, HRM requires n\_train $\geq$ 2500 to match LoRA; below that, LoRA's simpler parameterization is more data-efficient. 
At n\_train=10000 (paper default), HRM substantially outperforms LoRA. 
On parity task, both adapters are near-chance at all data sizes. As a result, the practical guidance from the ablation (for these tasks) is that HRM requires a minimum number of ($\sim$2000 in this case) training sequences to be competitive with LoRA on DFA-difficulty tasks.

%%%%%%%%%%%%%%%%%%%%%%%%%%%%%%%%%%%%%%%%%%%%%%%%%%%%%%%%%%%%%%%%%%%%%%%%%%%%%%%
%%%%%%%%%%%%%%%%%%%%%%%%%%%%%%%%%%%%%%%%%%%%%%%%%%%%%%%%%%%%%%%%%%%%%%%%%%%%%%%

\end{document}